\documentclass[sigconf]{acmart}

\usepackage{subcaption}

\AtBeginDocument{%
  }

\copyrightyear{2026}
\acmYear{2026}
\setcopyright{cc}
\setcctype{by}
\acmConference[CHI EA '26]{Extended Abstracts of the 2026 CHI Conference on Human Factors in Computing Systems}{April 13--17, 2026}{Barcelona, Spain}
\acmBooktitle{Extended Abstracts of the 2026 CHI Conference on Human Factors in Computing Systems (CHI EA '26), April 13--17, 2026, Barcelona, Spain}
\acmDOI{10.1145/3772363.3798695}
\acmISBN{979-8-4007-2281-3/2026/04}

\begin{document}

\title{ReMe: Scaffolding Personalized Cognitive Training via Controllable LLM-Mediated Conversations}


\author{Zilong Wang}
\authornote{These authors contributed equally to this work.}
\affiliation{%
  \institution{Microsoft Research}
  \city{Shanghai}
  \country{China}}
\email{wangzilong@microsoft.com}

\author{Nan Chen}
\authornotemark[1]
\affiliation{%
  \institution{Microsoft Research}
  \city{Shanghai}
  \country{China}}
\email{nanchen@microsoft.com}

\author{Luna K. Qiu}
\affiliation{%
  \institution{Microsoft Research}
  \city{Shanghai}
  \country{China}}
\email{lunaqiu@microsoft.com}

\author{Ling Yue}
\affiliation{%
  \institution{Department of Geriatric Psychiatry, Shanghai Mental Health Center,
Shanghai Jiao Tong University School of Medicine}
  \city{Shanghai}
  \country{China}}
\email{bellinthemoon@sjtu.edu.cn}

\author{Geli Guo}
\affiliation{%
  \institution{Microsoft Research}
  \city{Beijing}
  \country{China}}
\email{v-guobella@microsoft.com}

\author{Yang Ou}
\affiliation{%
  \institution{Microsoft Research}
  \city{Beijing}
  \country{China}}
\email{yang.ou@microsoft.com}

\author{Shiqi Jiang}
\affiliation{%
  \institution{Microsoft Research}
  \city{Shanghai}
  \country{China}}
\email{shijiang@microsoft.com}

\author{Yuqing Yang}
\affiliation{%
  \institution{Microsoft Research}
  \city{Shanghai}
  \country{China}}
\email{Yuqing.Yang@microsoft.com}

\author{Lili Qiu}
\affiliation{%
  \institution{Microsoft Research}
  \city{Shanghai}
  \country{China}}
\email{liliqiu@microsoft.com}


\begin{abstract}
Global aging calls for scalable and engaging cognitive interventions. Computerized cognitive training (CCT) is a promising non-pharmacological approach, yet many unsupervised programs rely on rigid, hand-authored puzzles that are difficult to personalize and can hinder adherence. Large language models (LLMs) offer more natural interaction, but their open-ended generation complicates the controlled task structure required for cognitive training.
We present ReMe, a web-based framework that scaffolds cognitive training through controllable LLM-mediated conversations, addressing both rigidity in conventional CCT content and the need for conversational controllability. ReMe features a modular Puzzle Engine that represents training activities as reusable puzzle groups specified by structured templates and constraint rules, enabling rapid development of dialogue-based word games and personalized tasks grounded in user context. By integrating personal life logs, ReMe supports Life Recall activities for episodic-memory practice through guided retrieval and progressive cues. A community pilot with 32 adults aged 50+ provides initial feasibility signals.
\end{abstract}



\begin{teaserfigure}
  \includegraphics[width=\textwidth]{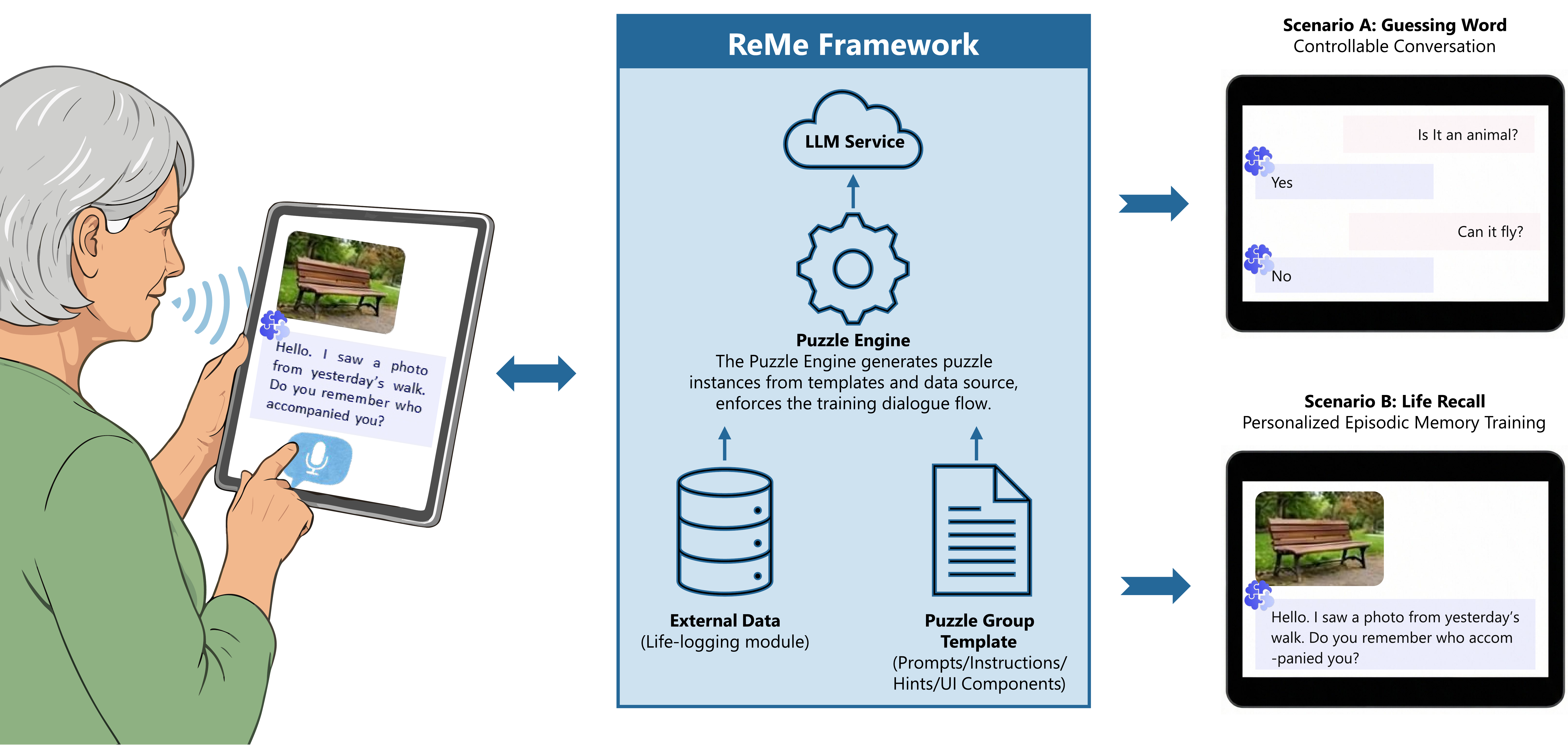}
  \caption{ReMe overview: Users interact with an LLM-powered voice-based chatbot through a multimodal training interface. The Puzzle Engine integrates information from life logs (and other data sources when needed) to create a puzzle instance and manage the conversational workflow.}
  \Description{Overview diagram of the ReMe framework for LLM-mediated cognitive training. On the left, an older adult user holds a tablet and interacts with a chatbot through a multimodal voice-first interface. The screen shows an image from a life log and a conversational prompt asking the user to recall details. In the center, the ReMe Framework diagram illustrates the system workflow: a Puzzle Engine connects to an LLM service, retrieves life logs and external data, and instantiates puzzle templates that specify prompts, instructions, hints, and task constraints. On the right, two example training scenarios show different puzzle types. Scenario A shows a guessing word game in which the chatbot answers yes or no questions. Scenario B shows a life-recall task in which the chatbot verifies recollections and provides hints and images. Arrows indicate the bidirectional flow of information between the user, the Puzzle Engine, and the instantiated puzzle scenarios.}
  \label{fig:teaser}
\end{teaserfigure}


\maketitle

\section{Introduction}
The proportion of older adults has increased substantially over recent decades~\cite{owid_unwpp_0_4_high_2024}, accompanied by a growing burden of cognitive disorders such as Alzheimer’s disease~\cite{alzheimers2024facts}. Currently, no treatment can reliably reverse Alzheimer’s~\cite{WU2025100112}; therefore, prevention and early intervention remain essential. In this context, cognitive training has emerged as a promising non-pharmacological strategy.

Computerized cognitive training (CCT) can improve certain cognitive functions in older adults ~\cite{kueider2012computerized,butler2018does}, including long-term benefits for domains such as reasoning and processing speed observed in the ACTIVE trial ~\cite{willis2006long,rebok2014ten}. However, real-world impact in unsupervised settings remains constrained by a design gap. Many scalable programs rely on rigid, hand-authored puzzles to preserve task structure, making content difficult to personalize. This rigidity can undermine engagement and adherence ~\cite{vermeir2020effects,li2024effects}. Moreover, effective interventions often still depend on professional supervision for sustained use and usability support, while unsupervised training tends to show smaller effects ~\cite{chan2024computerized}. This is particularly challenging for episodic-memory practice, where ecologically grounded tasks are difficult to deliver without supervision ~\cite{ranganath2011can,rudebeck2012potential,zimmermann2016transfer}.

LLM-powered voice-based chatbots create new opportunities for cognitive training. Compared with traditional interfaces, conversational interaction offers a low-barrier way to engage older adults, while LLMs provide open-world knowledge that can support more diverse tasks and enable personalization using user profiles, performance history, and everyday records ~\cite{brown2020language,achiam2023gpt}. LLM-based chatbots have also shown promise in medicine and mental health support, motivating careful exploration in cognitive health contexts ~\cite{lee2023benefits,balcombe2023ai,abd2019overview}. However, cognitive training relies on structured tasks, enforceable interaction constraints, and interpretable feedback to ensure consistent practice and track progress. Open-ended generation can drift off-task or break required formats, weakening training fidelity and progression. This creates a core tension for LLM-mediated cognitive training: how to preserve natural, engaging dialogue while maintaining the controllability needed for structured practice. ReMe addresses this tension by scaffolding cognitive training through controllable LLM-mediated conversations.


We present \textbf{ReMe}, a web-based framework for building LLM-powered chatbots for personalized cognitive training. ReMe is guided by four design goals: \textbf{(DG1) controllable dialogue-based training} that preserves task objectives while leveraging conversational engagement, drawing on scaffolding principles to maintain focus and reduce degrees of freedom~\cite{wood1976role,dhillon2024shaping}; \textbf{(DG2) personalized episodic-memory tasks} grounded in each user's life context for unsupervised practice, informed by the self-reference effect~\cite{symons1997self} and utilizing life logs as autobiographical memory cues~\cite{sellen2007life} (rather than "verifiable facts"); \textbf{(DG3) low-friction interaction for older adults} through voice-first and lightweight UI to accommodate age-related declines in fine motor skills~\cite{findlater2013age} and minimize extraneous cognitive load~\cite{liu2021mobile} via natural modalities~\cite{vollmer2017emerging}; and \textbf{(DG4) an extensible puzzle framework} for rapid creation and iteration to simplify LLM prompting complexities~\cite{zamfirescu2023johnny} and sustain adherence through task variety. Concretely, ReMe represents training activities as reusable \textit{puzzle groups} with explicit templates (prompts, instructions, hints, and interaction components) and enforceable constraint rules, and connects them with \textbf{personal life logs} that provide personally grounded real-life cues for guided episodic-memory practice (Fig.~\ref{fig:teaser}).

\noindent\textbf{Contributions:}
\begin{itemize}
    \item \textbf{ReMe}, a reusable framework for rapidly prototyping LLM-mediated cognitive training with modular task abstractions and reusable interaction components.
    \item A controlled-generation interaction paradigm that supports training-oriented conversations through enforceable constraints and structured task progression.
    \item Feasibility evidence from a public outreach pilot study with older adults (N=32, aged 50+) showing initial usability and engagement signals.
\end{itemize}

\section{System Design and Implementation}
ReMe centers around three components (Fig.~\ref{fig:teaser}). 
Guided by the design goals introduced above, the \textbf{Puzzle Engine} instantiates training tasks from reusable specifications and manages task-oriented conversational workflows. 
The \textbf{Life-Logging Module} supplies personal episodic materials for scenario-based memory training. 
The \textbf{Training User Interface} supports voice-first multimodal dialogue with reusable interaction components beyond conventional chat.

\subsection{Puzzle Engine: Puzzle Groups and Instance Creation}
ReMe organizes tasks into \textit{puzzle groups}, which are reusable task definitions that share objectives, instructions, and interaction rules. 
Each puzzle group includes a group name, a prompt template used for model inference, an instruction template shown to users, a hint template describing how hints are delivered (including multimodal hints when needed), and optional interaction components required by the task (e.g., rating, end-of-session, drawing). 
Puzzle groups also specify explicit interaction constraints, such as required response formats (e.g., yes/no-only answering), to preserve training fidelity (DG1).
At runtime, the engine creates a puzzle instance by populating templates with relevant data sourced from randomized generation, external sources, or life logs, prepares any hint payloads (e.g., a related life-log photo), and initiates a chat session. 
This template-based authoring enables rapid customization and iteration without implementing a new application for each task (DG4).

To balance natural dialogue with training controllability (DG1), ReMe encodes task constraints and recovery behaviors in the puzzle-group system prompt. 
The model is instructed to perform internal planning before producing a user-visible response, and to steer the dialogue back to valid interaction forms when violations occur. Currently, constraint enforcement is prompt-level: the system prompt specifies valid interaction patterns per puzzle group, and the model performs a ReAct-style reasoning step over the full session history each turn to detect violations and issue redirects before advancing the task. ~\cite{yao2022react}
For example, in a yes/no riddle task, if a user requests disallowed information (e.g., ``What is the first letter?''), the chatbot restates the constraint and redirects the user to a valid yes/no question about properties or usage.

\subsection{Life-Logging Module}
For episodic-memory training, ReMe includes a life-logging module where users can upload life log entries containing text and images. 
The module stores timestamps, descriptions, and images, and supports later retrieval for training. 
These life logs provide personally grounded content that can be transformed into individualized recall tasks (e.g., recalling a meal or an event from the previous day), enabling personalization in unsupervised practice (DG2). 
The retrieved logs also support progressive hinting with textual or image cues, allowing the system to provide assistance while keeping tasks grounded in real-life materials (DG2). To mitigate hallucination, the system prompt restricts factual claims to the retrieved log entry and instructs the model to explicitly ground each verification step in the retrieved artifacts before producing a response ~\cite{zhang2025hallucination}.

\subsection{Training User Interface and Reusable Components}
ReMe defines \textit{Chat Message} as the basic unit of conversation, containing voice, text, and images. 
Messages are model-independent and can be converted into the input format required by a chosen model, keeping puzzle definitions decoupled from any specific LLM (DG4). 
The UI supports voice-first interaction complemented by lightweight widgets to reduce operational burden for older adults (DG3). 
To support specialized puzzle interactions, ReMe provides reusable UI components invoked via tags and parameters in the chatbot output, such as: (i) \textit{hint} to display pre-configured multimodal messages, (ii) \textit{rating} to collect structured feedback, (iii) \textit{end} to terminate a session, and (iv) \textit{draw} to enable drawing input (e.g., a clock drawing task). 
These components externalize task control and structured input beyond free-form dialogue, making interactions consistent across puzzle types and easy to extend (DG4), while maintaining low-friction use in voice-based sessions (DG3).

\section{User Scenario}
We illustrate ReMe with two puzzle groups that cover constrained dialogue for structured training and life-log grounded episodic recall.

\begin{figure}[t]
    \centering
    \begin{subfigure}[t]{0.49\linewidth}
        \centering
        \includegraphics[width=\linewidth]{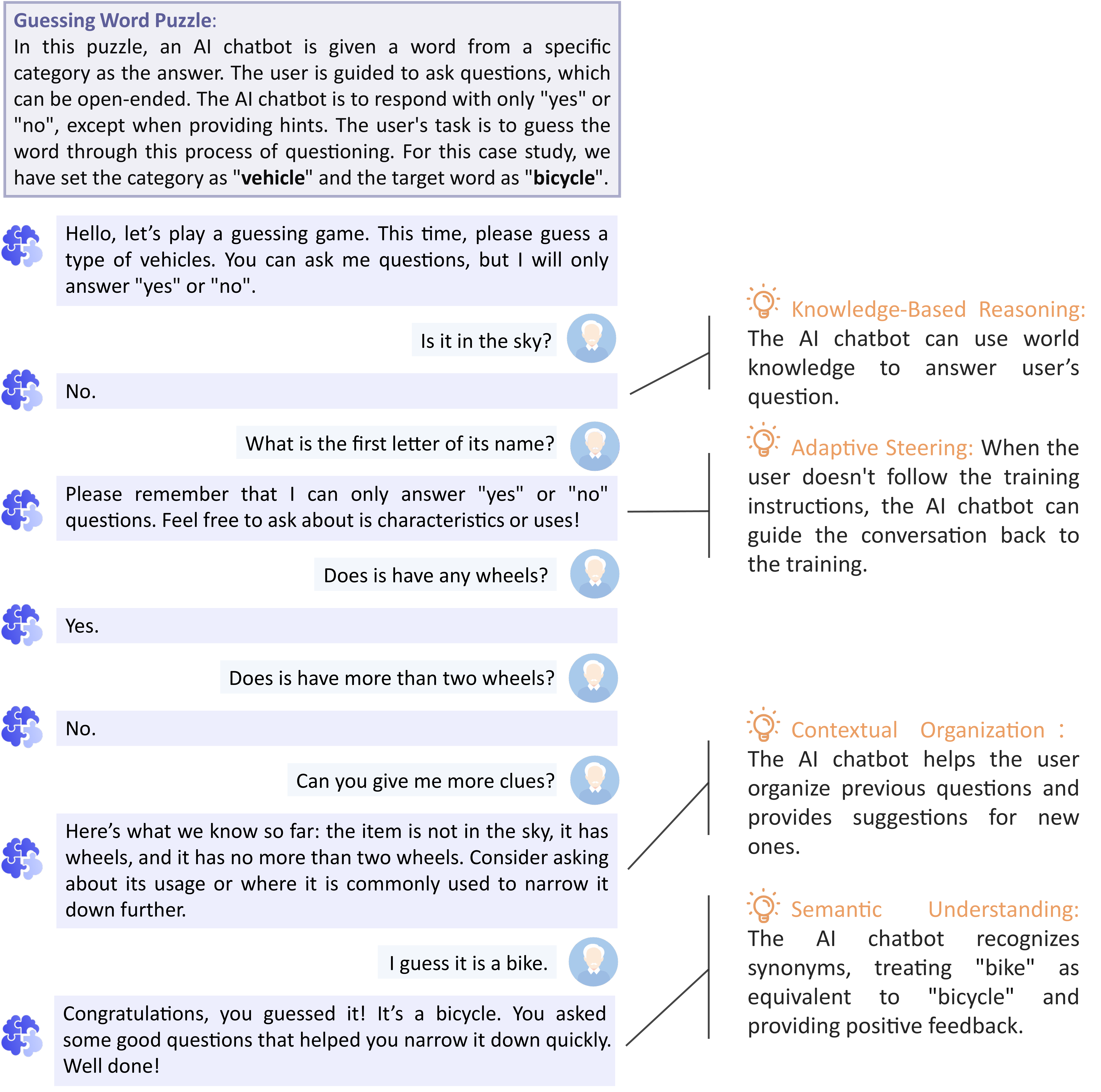}
        \caption{Guessing Word.}
        \label{fig:cases-a}
    \end{subfigure}\hfill
    \begin{subfigure}[t]{0.49\linewidth}
        \centering
        \includegraphics[width=\linewidth]{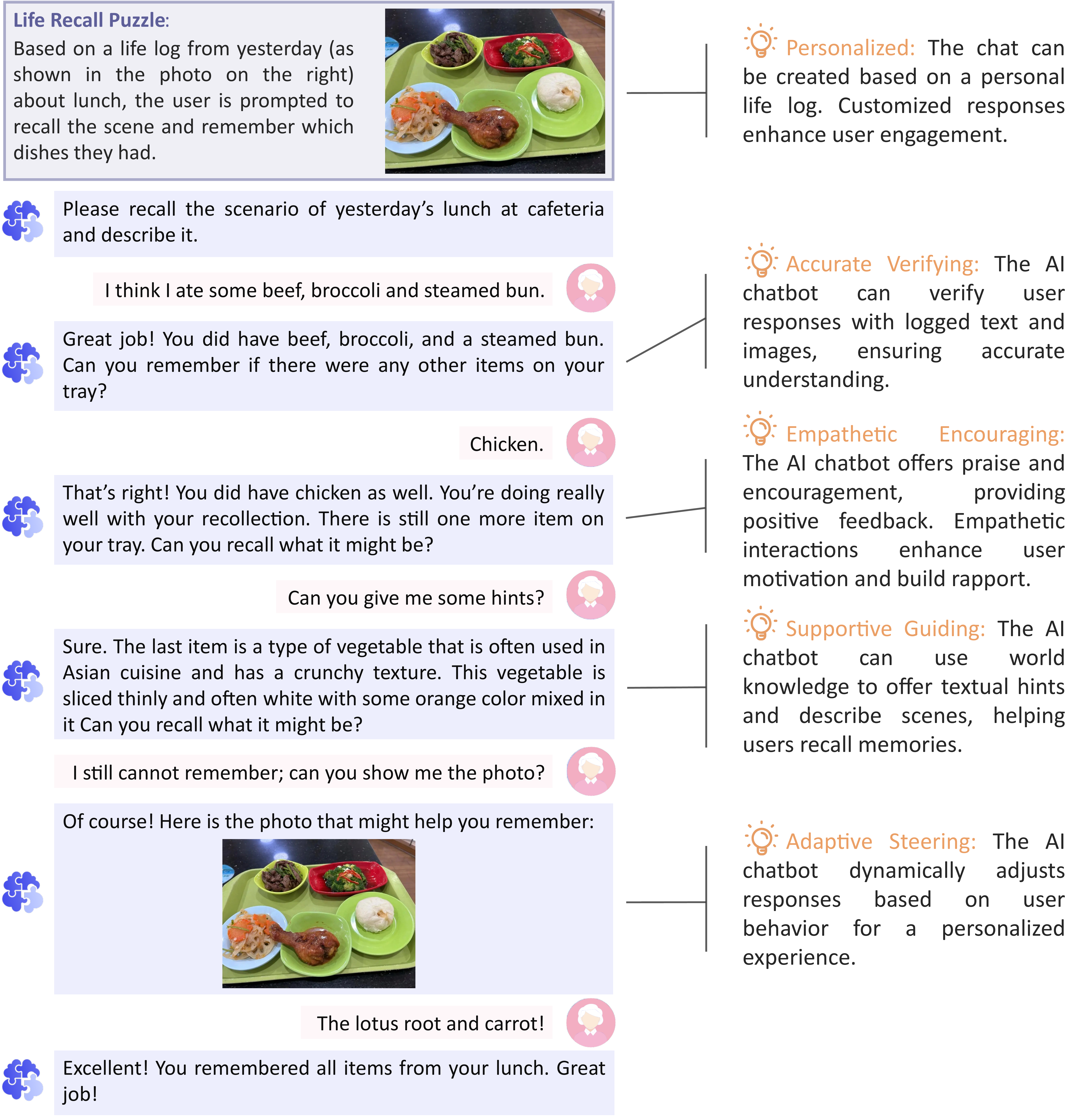}
        \caption{Life Recall.}
        \label{fig:cases-b}
    \end{subfigure}
    \caption{Training cases instantiated by ReMe.}
    \label{fig:cases}
    \Description{Two example puzzle sessions generated by the ReMe framework. The left panel shows a Guessing Word puzzle. The chatbot provides a category and answers only yes or no questions while the user attempts to guess a hidden target word. The interface shows sequential chat turns, constraint reminders, hint summaries, and acceptance of a correct guess. The right panel shows a Life Recall puzzle based on a life-log entry. The chatbot prompts the user to recall items from a lunch gathering recorded the previous day. The system verifies recalled details, provides progressively more specific textual hints, and optionally reveals a photo from the life log to support retrieval. The surrounding annotations highlight the capabilities the system employs during the tasks, such as knowledge-based reasoning, adaptive steering, contextual organization, accurate verification, and empathetic guidance.}
\end{figure}

\subsection{Persona and Context}
Consider an older adult user engaging in short daily training sessions at home, primarily through voice with lightweight UI support when needed (DG3).

\subsection{Scenario A: Guessing Word (Open-World Yes/No Riddle)}
\paragraph{Goal.} Train reasoning and language through an open-world guessing game under a strict yes/no answering constraint (DG1). This stepwise, information-seeking interaction is intended to engage working memory, as the user must maintain and update partial hypotheses across turns while actively drawing on and refining everyday language skills and commonsense world knowledge.
\paragraph{Workflow.} The user is given a category and asks a sequence of yes/no questions to narrow down a hidden target. The chatbot answers in yes/no form, enforces the format, and may summarize confirmed facts to guide subsequent questions.
\paragraph{Mechanism.} Each session is defined by a category and a target word. The puzzle group encodes knowledge-based yes/no responding, violation steering for invalid queries, brief fact summarization, and acceptance of equivalent guesses.
\paragraph{Example.} In Fig.~\ref{fig:cases}a, the category is \textit{vehicle} and the target is \textit{bicycle}. After a valid question (``Is it in the sky?'') receives ``no,'' the user asks an invalid request (``What is the first letter?''). The chatbot restates the constraint, redirects the user to ask a valid yes/no question, summarizes known facts, and accepts ``bike'' as equivalent to \textit{bicycle}.

\subsection{Scenario B: Life Recall (Episodic Memory)}
\paragraph{Goal.} Prompt users to recall details of a recent life event recorded in a life log, with progressive cues when needed (DG2). In doing so, the task supports individualized training of episodic memory grounded in the user's own autobiographical history.
\paragraph{Workflow.} The user first records daily events as life logs. During training, the system retrieves a relevant entry and prompts the user to recall details. The chatbot verifies responses and provides progressively stronger hints, including an optional photo cue.
\paragraph{Mechanism.} The puzzle instance is generated from a retrieved life-log entry. The chatbot guides recall with structured prompts, verifies recalled content against logged text and images, and adapts hint specificity based on user progress.
\paragraph{Example.} In Fig.~\ref{fig:cases}b, the system retrieves a logged ``yesterday's lunch'' entry and prompts recall. The chatbot confirms recalled items, provides increasingly specific textual hints when recall is incomplete, and can reveal the logged photo to support retrieval.

\subsection{What the Scenarios Demonstrate}
Both puzzles are authored as puzzle groups that reuse templates, constraints, and UI components, enabling rapid iteration without implementing task-specific interfaces from scratch (DG4).

\section{Pilot Study: Usability and Engagement Signals}
We conducted a community outreach pilot study to assess feasibility signals of ReMe in a low-support setting, focusing on usability and engagement rather than cognitive efficacy. 
The pilot was conducted on World Alzheimer's Day following a public awareness lecture. 
After a brief introduction, participants voluntarily tried both puzzles on designated devices and completed a questionnaire.

\textbf{Ethics and privacy.}
The pilot study was approved by the Institutional Review Board (IRB). To minimize privacy risks, the Life Recall puzzle used a pre-set image as a case study rather than participants' personal life logs. As a result, the pilot is limited to evaluating the interaction flow and usability of Life Recall with generic content; testing with participants' own life-log data remains an important next step.
No identifiable personal information was collected or retained, and all session data were cleared after the session concluded.

\textbf{Participants.}
We report questionnaire results from 32 participants aged 50 years or older (n=32).
Age distribution was: 50--60 (n=9, 28.13\%), 60--70 (n=15, 46.88\%), and 70+ (n=8, 25.00\%).
Gender distribution was: female (n=23, 71.88\%) and male (n=9, 28.13\%).
Within the same analysis sample, 84.38\% reported at least a high-school education, 37.50\% reported taking daily medication for chronic conditions, and 75.00\% reported using a mobile phone for more than 2 hours per day.
In addition, 25.00\% reported prior experience with cognitive training software, and half of these participants reported that the interfaces of previous tools were unfriendly.

\textbf{Measures.}
Participants rated perceived difficulty and enjoyment for each puzzle using 5-point Likert scales (difficulty: 1=easiest, 5=hardest; enjoyment: 1=least enjoyable, 5=most enjoyable).

\textbf{Results.}
Both puzzles were rated as moderately difficult and enjoyable.
For \textit{Guessing Word}, difficulty was median=3 (IQR=2--3) and enjoyment was median=3 (IQR=2--4); 81.25\% rated difficulty as 2--4/5 and 71.88\% rated enjoyment as $\geq$3/5.
Participants explicitly highlighted the benefits of the voice-first interface compared to touching screens: ``\textit{This(ReMe) feels flexible because I can speak directly. Previous training tools required interaction with the phone, which was difficult for me; this is much easier.}''

For \textit{Life Recall}, difficulty was median=3 (IQR=2--3) and enjoyment was median=3 (IQR=3--4); 81.25\% rated difficulty as 2--4/5 and 75.00\% rated enjoyment as $\geq$3/5.
Although the pilot used generic images, participants expressed a strong desire for personalization: ``\textit{The recall task is interesting, it would be more engaging if the content were based on my own life experiences when training.}''

For longer-term willingness, 31.25\% reported being willing to invest at least 30 minutes per day to reduce Alzheimer's risk, and 90.62\% indicated they would likely continue training for at least one month if recommended.

\section{Discussions, Limitations, and Risks}
The pilot suggests that ReMe-based puzzles can be usable and engaging in a low-support setting. Together with our two case studies, these results indicate ReMe's potential to deliver both episodic-memory practice grounded in everyday records and open-world reasoning and language training under explicit interaction constraints. More rigorous studies are needed to evaluate cognitive outcomes.

\textbf{Design implications.} ReMe illustrates a controllable conversational training paradigm that preserves task fidelity through explicit constraints and recovery behaviors, while retaining low-barrier dialogue. It also demonstrates a path from life logs to ecologically grounded episodic-memory practice via retrieval-based prompting and progressive cues. As a reusable framework, ReMe lowers the cost of authoring and iterating training tasks through puzzle-group templates and reusable UI components.

\textbf{Limitations.} Our evidence is limited to short-term feasibility signals and does not establish efficacy or transfer. Voice-first, turn-based interaction may still feel less natural than human conversation and can be sensitive to noisy environments, requiring onboarding for some users. High-quality interaction depends on capable LLMs, introducing cost and latency. Personalized recall quality also depends on life-log coverage.

\textbf{Risks and mitigation directions.} Life-log based personalization raises privacy and security risks; mitigations include data minimization, user controls, and secure storage~\cite{ali2019insight,wilkowska2020two}. LLM hallucinations may harm safety and training validity; mitigations include grounding in retrieved artifacts, conservative prompting, and output filtering~\cite{zhang2025siren,anh2025survey,asgari2025framework}. Prolonged empathetic interaction may increase over-reliance; mitigations include transparency, pacing or usage limits, and integration with real-world support.

\textbf{Future work.} We will expand ReMe with additional puzzle groups and difficulty adaptation, and conduct longitudinal studies with validated cognitive assessments. We will further optimize end-to-end responsiveness for voice-first interaction and investigate privacy-preserving approaches for handling life-log data in real-world deployments.

\bibliographystyle{ACM-Reference-Format}
\bibliography{sample-base}



\end{document}